\documentclass[runningheads]{llncs}
\usepackage{graphicx}
\usepackage{comment}
\usepackage{amsmath,amssymb} 
\usepackage{color}
\usepackage{bm}

\usepackage{hhline}
\usepackage[normalem]{ulem}
\usepackage[toc,page,title]{appendix}

\newcommand{\EPos}{\bm{c}}
\newcommand{\EDir}{\bm{\omega}_o}
\newcommand{\LDir}{\bm{\omega}_i}
\newcommand{\LPos}{\bm{l}}
\newcommand{\HDir}{\bm{h}}

\newcommand{\PosX}{\bm{x}}
\newcommand{\posx}{x}
\newcommand{\SNormal}{\bm{n}}
\newcommand{\SReflect}{R}
\newcommand{\Radiance}{L}
\newcommand{\RadiancePerM}{L}
\newcommand{\accumAlpha}{\alpha}
\newcommand{\cs}{c\rightarrow s}
\newcommand{\csp}{c\rightarrow(s-1)}
\newcommand{\lt}{l\rightarrow t}
\newcommand{\ltp}{l\rightarrow(t-1)}

\newcommand{\Extinct}{\sigma_t}

\newcommand{\Trans}{\tau}

\newcommand{\SPony}{\textsc{Pony}}
\newcommand{\SGirl}{\textsc{Girl}}
\newcommand{\SHouse}{\textsc{House}}

\newcommand{\SDisney}{\textsc{Disney}}
\newcommand{\SLion}{\textsc{Animals}}
\newcommand{\SCaptain}{\textsc{Captain}}

\newcommand{\boldstart}[1]{\noindent\textbf{#1}}
\newcommand{\boldstartspace}[1]{\vspace{0.1in}\noindent\textbf{#1}}
\newcommand{\Comment}[1]{}

\newcommand{\SAlbedo}{\bm{a}}
\newcommand{\SRough}{\gamma}

\begin{document}
\pagestyle{headings}
\mainmatter
\def\ECCVSubNumber{362}  

\title{Deep Reflectance Volumes: \\ Relightable Reconstructions from  Multi-View \\Photometric Images} 

\titlerunning{Deep Reflectance Volumes}

%
\author{
    Sai Bi\textsuperscript{1}, 
    Zexiang Xu\textsuperscript{1,2}, 
    Kalyan Sunkavalli\textsuperscript{2},
    Milo\v{s} Ha\v{s}an\textsuperscript{2},
    Yannick Hold-Geoffroy\textsuperscript{2},
    David Kriegman\textsuperscript{1},
    Ravi Ramamoorthi\textsuperscript{1}
}
%
\authorrunning{Bi et al.}
%

\institute{University of California, San Diego \and
Adobe Research
}


\maketitle

\begin{abstract}

We present a deep learning approach to reconstruct scene appearance from unstructured images captured under collocated point lighting. At the heart of Deep Reflectance Volumes is a novel volumetric scene representation consisting of opacity, surface normal and reflectance voxel grids. We present a novel physically-based differentiable volume ray marching framework to render these scene volumes under arbitrary viewpoint and lighting. This allows us to optimize the scene volumes to minimize the error between their rendered images  and the captured images. Our method is able to reconstruct real scenes with challenging non-Lambertian reflectance and complex geometry with occlusions and shadowing. Moreover, it accurately generalizes to \emph{novel} viewpoints and lighting, including non-collocated lighting, rendering photorealistic images that are significantly better than  state-of-the-art mesh-based methods. We also show that our learned reflectance volumes are editable, allowing for modifying the materials of the captured scenes. 

\keywords{View synthesis, relighting, appearance acquisition, neural rendering}
\end{abstract}

\section{Introduction}\label{sec:intro}

Capturing a real scene and re-rendering it under novel lighting conditions and viewpoints
is one of the core challenges in computer vision and graphics.
This is classically done by reconstructing the 3D scene geometry, typically in the form of a mesh, and computing per-vertex colors or reflectance parameters, to support arbitrary re-rendering. 
However, 
3D reconstruction methods like multi-view stereo are prone to errors in textureless and non-Lambertian regions \cite{nam2018practical,schoenberger2016mvs}, 
and accurate reflectance acquisition usually requires dense, calibrated capture using  
sophisticated devices \cite{baek2018simultaneous,xia2016recovering}.

Recent works have proposed learning-based approaches to capture scene appearance.
One class of methods use surface-based representations \cite{groueix2018papier,kanazawa2018learning} but are restricted to specific scene categories and cannot synthesize photo-realistic images. 
Other methods bypass explicit reconstruction,
instead focusing on relighting \cite{xu2018deep} or view synthesis sub-problems \cite{lombardi2019neural,xu2019deep}.


\begin{figure}[t]

\includegraphics[width=\textwidth]{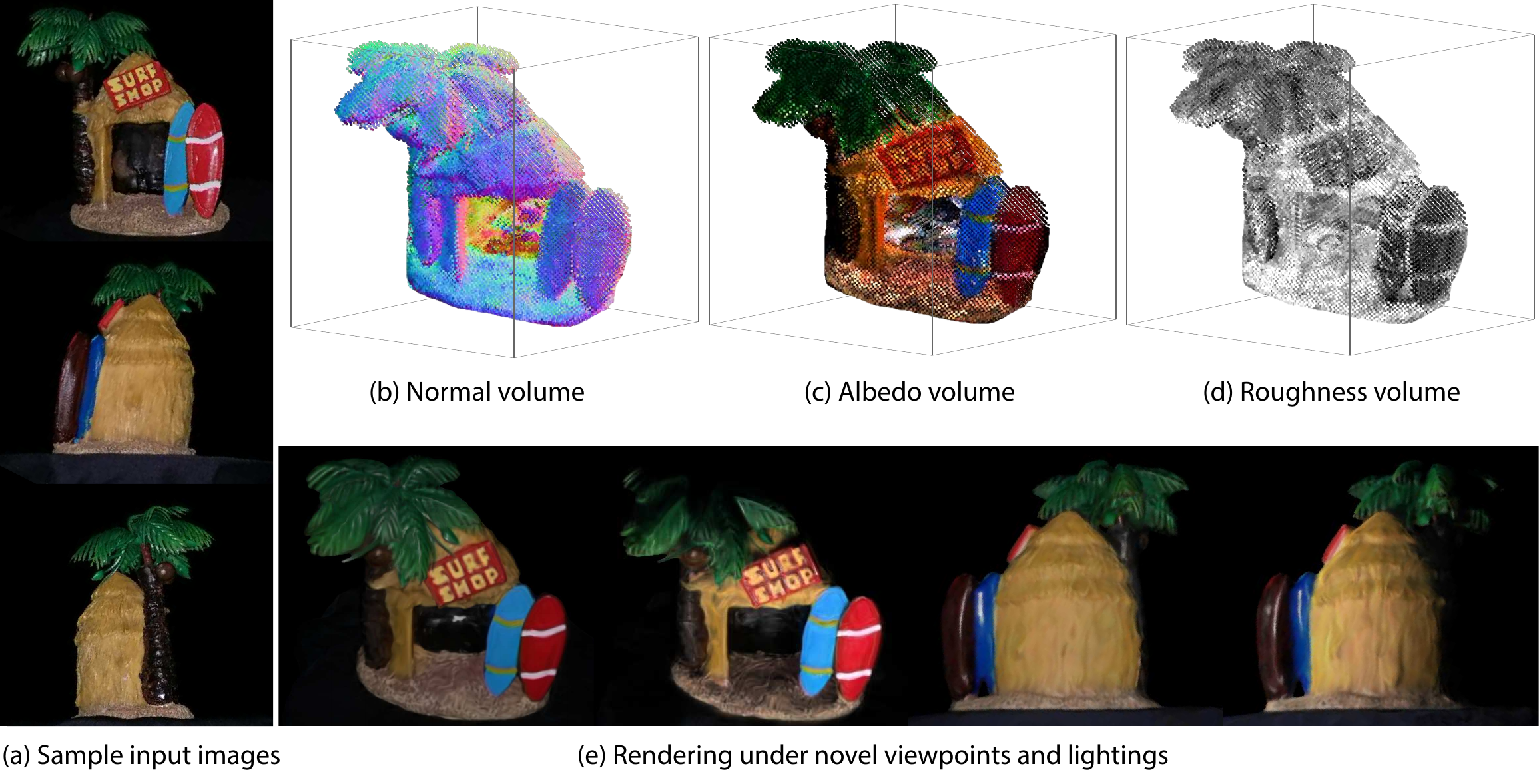}
\caption{Given a set of images taken using a  mobile phone with flashlight (sampled images are shown in (a)), our method learns a volume representation of the captured object by estimating the opacity volume, normal volume (b) and reflectance volumes such as albedo (c) and roughness (d). Our volume representation enables free navigation of the object under 
arbitrary viewpoints and novel lighting conditions (e).}
    \label{fig:teaser}
\end{figure}


Our goal is to make high-quality scene acquisition and rendering practical with off-the-shelf devices under mildly controlled conditions.
We use a set of unstructured images captured around a scene by a single mobile phone camera with flash illumination in a dark room.
This practical setup acquires multi-view images under collocated viewing and lighting directions---referred to as photometric images \cite{xu2019deep}.
While the high-frequency appearance variation in these images (due to sharp specular highlights and shadows) can result in
low-quality mesh reconstruction 
from state-of-the-art methods (see Fig.~\ref{fig:meshbased}),
we show that our method can accurately model the scene and realistically reproduce complex appearance information like specularities and occlusions.

At the heart of our method is a novel, physically-based neural volume rendering framework.
We train a deep neural network that simultaneously learns the geometry and \emph{reflectance} of a scene as volumes.
We leverage a decoder-like network architecture,
where an encoding vector together with the corresponding network parameters are learned during a per-scene optimization (training) process.
Our network decodes a volumetric scene representation
consisting of opacity, normal, diffuse color and roughness volumes, which model the global geometry,
local surface orientations and spatially-varying reflectance parameters of the scene, respectively.
These volumes are supplied to a differentiable rendering module
to render images with collocated light-view settings at training time, and arbitrary light-view settings at inference time (see Fig.~\ref{fig:raymarching}).

We base our differentiable rendering module on classical volume ray marching approaches
with opacity (alpha) accumulation and compositing \cite{kniss2003model,wittenbrink1998opacity}.
In particular, we compute point-wise shading using local normal and reflectance properties,
and accumulate the shaded colors with opacities along each marching ray of sight.
Unlike the opacity used in previous view synthesis work~\cite{lombardi2019neural,zhou2018stereo}
that is only accumulated along view directions,
we propose to learn global scene opacity that can be accumulated from both view and light directions.
As shown in Fig.~\ref{fig:teaser}, we demonstrate that our scene opacity can be effectively learned
and used to compute accurate hard shadows under \emph{novel lighting}, despite the fact
that the training process never observed images with shadows that are taken under non-collocated view-light setups.
Moreover, different from previous volume-based works~\cite{lombardi2019neural,zhou2018stereo} that learn a single color at each voxel,
we reconstruct per-voxel reflectance and handle complex materials with high glossiness.  
Our neural rendering framework thus enables
rendering with complex view-dependent and light-dependent shading effects including specularities, occlusions and shadows.
    We compare against a state-of-the-art mesh-based method~\cite{nam2018practical}, and demonstrate
    that our method is able to achieve more accurate reconstructions and renderings (see Fig.~\ref{fig:meshbased}).  
    We also show that our approach supports scene material editing by modifying the reconstructed reflectance volumes (see Fig.~\ref{fig:material}).
    To summarize, our contributions are:
\begin{enumerate}
    \item[$-$] A practical neural rendering framework that reproduces
    high-quality geometry and appearance from unstructured mobile phone flash
    images and enables view synthesis, relighting, and scene editing.
    \item[$-$] A novel scene appearance representation using opacity, normal and reflectance volumes. 
    \item[$-$] A physically-based differentiable volume rendering 
    approach based on deep priors that can effectively 
    reconstruct the volumes from input flash images.
\end{enumerate}

\section{Related Works}\label{sec:related}
\boldstart{Geometry reconstruction.}
There is a long history in reconstructing 3D geometry from images 
using traditional structure from motion and 
multi-view stereo (MVS) 
pipelines~\cite{furukawa2009accurate,kutulakos2000theory,schoenberger2016mvs}.
Recently deep learning techniques have also been applied to 3D reconstruction with various representations, 
including volumes \cite{ji2017surfacenet,richter2018matryoshka}, point clouds \cite{achlioptas2018learning,paschalidou2018raynet,wang2018mvpnet}, depth maps \cite{huang2018deepmvs,yao2018mvsnet} and implicit functions \cite{chen2018learning,mescheder2018occupancy,niemeyer2020differentiable}.
We aim to model scene geometry for realistic image synthesis, 
for which mesh-based reconstruction \cite{kazhdan2006poisson,lorensen1987marching,Newcombe2011kinectFusion} is 
the most common way in many applications \cite{bi2017patch,nam2018practical,philip2019multi,zhou2014color}.
However, it remains challenging to reconstruct
    accurate meshes for challenging scenes where there 
    are textureless regions and thin structures,
and it is hard to incorporate a mesh into a deep 
learning framework \cite{ladicky2017point,liao2018deep};
the few mesh-based deep learning works \cite{groueix2018papier,kanazawa2018learning} 
are limited to category-specific reconstruction 
and cannot produce photo-realistic results.
Instead, we leverage a physically-based opacity 
volume representation that can be easily
embedded in a deep learning system to express scene geometry of arbitrary shapes. 


\boldstartspace{Reflectance acquisition.}
Reflectance of real materials is classically measured using sophisticated devices to 
densely acquire light-view samples \cite{foo1997gonioreflectometer,matusik2003data}, which is impractical for common users.
Recent works have improved the practicality with fewer samples \cite{nielsen2015optimal,xu2016minimal} and 
more practical devices (mobile phones) \cite{aittala2016neural,aittala2015two,hui2017reflectance,li2018materials}; 
however, most of them focus on flat planar objects.
A few single-view techniques based on photometric stereo \cite{alldrin2008photometric,goldman2009shape} or deep learning \cite{li2018learning} 
are able to handle arbitrary shape, 
but they merely recover limited single-view scene content.
To recover complete shape with spatially varying BRDF from multi-view inputs, 
previous works usually rely on a pre-reconstructed initial mesh and images captured under complex controlled setups 
to reconstruct per-vertex BRDFs
\cite{bi2020deep,kang2019learning,wu2015simultaneous,xia2016recovering,zhou2016sparse}.
While a recent work \cite{nam2018practical} uses a mobile phone for practical acquisition like ours, 
it still requires MVS-based mesh reconstruction, 
which is ineffective for challenging scenes with textureless, 
    specular and thin-structure regions.
In contrast, we reconstruct spatially varying volumetric reflectance via deep network based optimization;
we avoid using any initial geometry and propose to jointly reconstruct geometry and reflectance in a holistic framework.

\boldstartspace{Relighting and view synthesis.}
Image-based techniques have been extensively explored in graphics and vision 
to synthesize images under novel lighting and viewpoint without explicit complete reconstruction 
\cite{buehler2001unstructured,debevec2000acquiring,levoy1996light,peers2009compressive}.
Recently, deep learning has been applied to view synthesis 
and most methods leverage either view-dependent volumes 
\cite{srinivasan2019pushing,xu2019deep,zhou2018stereo} 
or canonical world-space volumes \cite{lombardi2019neural,sitzmann2019deepvoxels} for geometric-aware appearance inference. We extend them to a more general physically-based volumetric representation 
which explicitly expresses both geometry and reflectance, and enables relighting with view synthesis.
On the other hand, learning-based relighting techniques have also been developed.
Purely image-based methods are able to relight scenes with realistic specularities and soft shadows from sparse inputs, 
but unable to reproduce accurate hard shadows \cite{kanamori2018relighting,sun2019single,xu2018deep,zhou2019deep};
some other methods~\cite{chen2020neural,philip2019multi} propose geometry-aware networks and make use of pre-acquired meshes for relighting and view synthesis, and
their performance 
is limited by the mesh reconstruction quality. 
A work~\cite{mildenhall2020nerf} concurrent to ours models scene geometry and appearance
by reconstructing a continuous radiance field for pure view synthesis. 
In contrast,
Deep Reflectance Volumes explicitly express scene geometry and reflectance, and reproduce accurate high-frequency specularities and hard shadows.
Ours is the first comprehensive neural rendering framework that enables both relighting and view synthesis
with complex shading effects.

\section{Rendering with Deep Reflectance Volumes}\label{sec:volume}
\begin{figure}[t]
    \includegraphics[width=\textwidth]{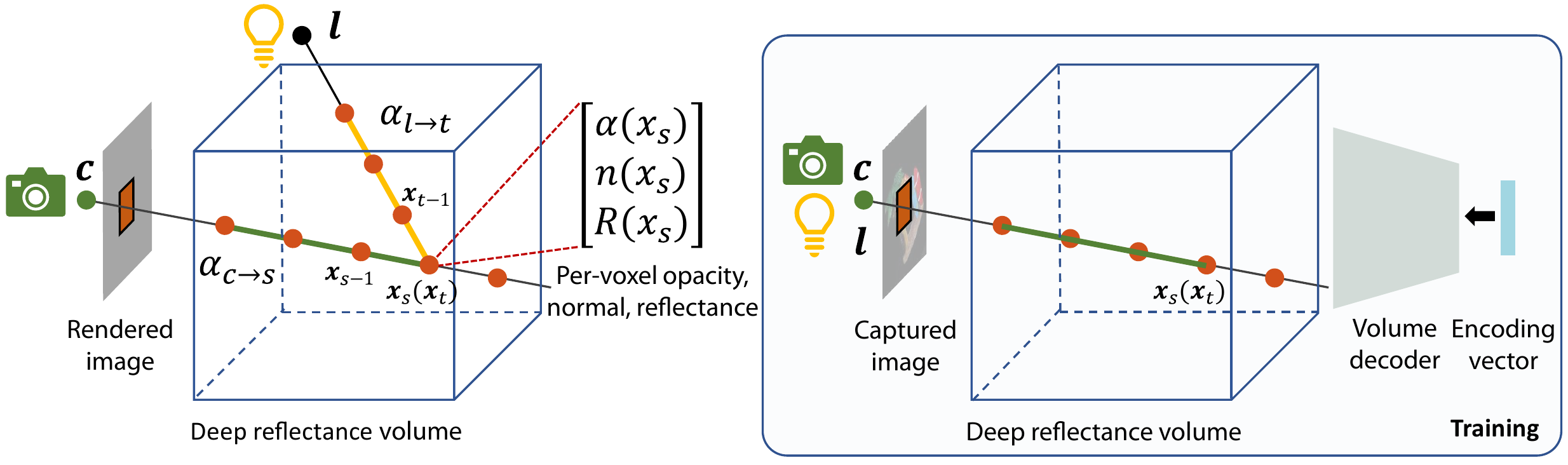}
    \caption{
           We propose Deep Reflectance Volume representation to capture scene geometry and 
           appearance, where each voxel consists of opacity $\alpha$, normal $n$ and reflectance (material coefficients) $R$. 
During rendering, we perform ray marching through each pixel and accumulate 
contributions from each point $\PosX_s$ along the ray. Each contribution
is calculated using the local normal, reflectance and lighting information.
We accumulate opacity from both the camera $\accumAlpha_{\cs}$ 
and the light $\accumAlpha_{\lt}$ to model the light transport loss in both occlusions and shadows. 
To predict such a volume, we 
start from an encoding vector, and decode it into a volume using a 3D convolutional
neural network; thus the combination of the encoding vector and network weights is the unknown variable being optimized (trained). We train on images captured with collocated
camera and light by enforcing a loss function between rendered
images and training images.
}
    \label{fig:raymarching}
\end{figure}

Unlike a mesh that is comprised of points with complex connectivity, a volume is a regular 3D grid, suitable for convolutional operations.
Volumes have been widely used in deep learning frameworks for 3D applications \cite{wu20153d,yao2018mvsnet}.
However, previous neural volumetric representations have only represented
pixel colors; this can be used for view synthesis \cite{lombardi2019neural,zhou2018stereo},
but does not support relighting or scene editing.
Instead, we propose to jointly learn geometry and reflectance (i.e. material parameters) volumes to enable broader rendering applications including view synthesis, relighting and material editing in a comprehensive framework.
Deep Reflectance Volumes are learned from a deep network and used to render images in a fully differentiable end-to-end process
as shown in Fig.~\ref{fig:raymarching}.
This is made possible by a new differentiable volume ray marching module, which is motivated by physically-based volume rendering.
In this section, we introduce our volume rendering method and volumetric scene representation.
We discuss how we learn these volumes from unstructured images in Sec.~\ref{sec:learning}.

\subsection{Volume rendering overview}

In general, volume rendering is governed by the physically-based volume rendering equation (radiative transfer equation) that describes
the radiance that arrives at a camera \cite{max1995optical,novak2018monte}:
\begin{equation}
    \Radiance(\EPos, \EDir) =  \int_0^\infty \Trans(\EPos,\PosX) [\RadiancePerM_e(\PosX,\EDir) + \RadiancePerM_s(\PosX,\EDir)] d\posx,
    \label{eqn:render}
\end{equation}
This equation integrates emitted, $\RadiancePerM_e$, and in-scattered, $\RadiancePerM_s$, light contributions along the ray starting at camera position $\EPos$ in the direction $-\EDir$.
Here, $\posx$ represents distance along the ray, and $\PosX=\EPos-\posx\EDir$ is the corresponding 3D point.
$\Trans(\EPos,\PosX)$ is the transmittance factor that governs the loss of light
 along the line segment between $\EPos$ and $\PosX$:
\begin{equation}
    \Trans(\EPos,\PosX) =  e^{-\int_0^\posx \Extinct(z) dz},
    \label{eqn:trans}
\end{equation}
where $\Extinct(z)$ is the extinction coefficient at location $z$ on the segment.
The in-scattered contribution is defined as:
\begin{equation}
    \RadiancePerM_s(\PosX,\EDir) =  \int_{\mathcal{S}} f_p(\PosX,\EDir,\LDir) \Radiance_i(\PosX,\LDir) d\LDir,
    \label{eqn:phaseInt}
\end{equation}
in which $\mathcal{S}$ is a unit sphere, $f_p(\PosX,\EDir,\LDir)$ is a generalized (unnormalized) phase function that expresses how light scatters at a point in the volume, and $\Radiance_i(\PosX,\LDir)$ is the incoming radiance that arrives at $\PosX$ from direction $\LDir$.

In theory, fully computing $\Radiance(\EPos, \EDir)$ requires multiple-scattering computation using Monte Carlo methods \cite{novak2018monte}, which is computationally expensive and unsuitable for deep learning techniques.
We consider a simplified case with a single point light, single scattering and no volumetric emission. The transmittance between the scattering location and the point light is handled the same way as between the scattering location and camera.
The generalized phase function $f_p(\PosX,\EDir,\LDir)$ becomes a reflectance function $f_r(\EDir,\LDir,\SNormal(\PosX),\SReflect(\PosX))$ which computes reflected radiance at $\PosX$ using its local surface normal $\SNormal(\PosX)$ and the reflectance parameters $\SReflect(\PosX)$ of a given surface reflectance model.
Therefore, Eqn.~\ref{eqn:render} and Eqn.~\ref{eqn:phaseInt} can be simplified and written concisely as \cite{kniss2003model,max1995optical}:
\begin{equation}
    \Radiance(\EPos, \EDir) =  \int_0^\infty \Trans(\EPos,\PosX) \Trans(\PosX,\LPos) f_r(\EDir,\LDir,\SNormal(\PosX),\SReflect(\PosX)) \Radiance_{\LPos}(\PosX,\LDir)  d\posx,
    \label{eqn:rendersimp}
\end{equation}
where $\LPos$ is the light position, $\LDir$ corresponds to the direction from $\PosX$ to $\LPos$, $\Trans(\EPos,\PosX)$ still represents the transmittance from the scattering point $\PosX$ to the camera $\EPos$, the term $\Trans(\PosX,\LPos)$ (that was implicitly involved in Eqn.~\ref{eqn:phaseInt})
is the transmittance from the light $\LPos$ to $\PosX$ and expresses light extinction before scattering,
and $\Radiance_{\LPos}(\PosX,\LDir)$ represents the light intensity arriving at $\PosX$ without considering light extinction.
\Comment{
We present deep reflectance volume representation 
that models both the scene geometry and appearance,
where each voxel consists of opacity $\alpha$, normal $n$ and reflectance (material coefficients) $R$.
We introduce a novel physically based volume ray marching module to render the volume representation.
In particular, a ray is marched through each pixel and
the per-step contributions from individual points $\PosX_s$ along the ray are accumulated. 
The contribution at each point
is calculated using the local normal and reflectance sampled from the volume and the lighting information.
We accumulate opacity from both the camera $\accumAlpha_{\cs}$ 
and the light $\accumAlpha_{\lt}$ to model the light transport loss in both occlusions and shadows.
We present a deep network to predict our volume representation.
The network starts from an encoding vector, and then decodes the vector to the volume using 3D convolutional layers; both the encoding vector and other network weights are optimized (trained). We train on images captured with collocated
camera and light by applying a loss function that minimizes the difference between rendered
images and captured images.
}

\subsection{A discretized, differentiable volume rendering module}

To make volume rendering practical in a learning framework, we further approximate Eqn.~\ref{eqn:rendersimp} by turning it into a discretized version,
which can be evaluated by ray marching
\cite{kniss2003model,max1995optical,wittenbrink1998opacity}.
This is classically expressed using opacity compositing, where opacity $\alpha$ is used to represent the transmittance with fixed ray marching step size $\Delta x$. 
Points are sequentially sampled along a given ray, $\EDir$ from the camera position, $\EPos$ as: 
\begin{align}
    \PosX_s = \PosX_{s-1} -\EDir \Delta x = \EPos - s\EDir \Delta x. \label{eqn:xsacc}
\end{align}
The radiance $\Radiance_s$ and opacity $\accumAlpha_{\cs}$ along this path, $\cs$, are recursively accumulated until $\PosX_s$ exits the volume as:
\begin{align}
    \Radiance_s & = \Radiance_{s-1} + [1-\accumAlpha_{\csp}][1-\accumAlpha_{\ltp}]\alpha(\PosX_s)\Radiance(\PosX_s),\label{eqn:racc}\\
    \accumAlpha_{\cs} & = \accumAlpha_{\csp} + [1-\accumAlpha_{\csp}]\alpha(\PosX_s),\label{eqn:acacc}\\
    \Radiance(\PosX_s) & = f_r(\EDir,\LDir,\SNormal(\PosX_s),\SReflect(\PosX_s)) \Radiance_{\LPos}(\PosX_s,\LDir). \label{eqn:shading}
\end{align}
Here, 
$\Radiance(\PosX_s)$ computes the reflected radiance from the reflectance function and the incoming light, $\accumAlpha_{\cs}$ represents the accumulated opacity from the camera $\EPos$ to point $\PosX_s$, and corresponds to $\Trans(\EPos,\PosX)$ in Eqn~\ref{eqn:rendersimp}.
$\accumAlpha_{\lt}$ represents the accumulated opacity from the light $\LPos$---i.e., $\Trans(\PosX,\LPos)$ in Eqn.~\ref{eqn:rendersimp}---and requires a \emph{separate} accumulation process over samples along the $\LPos\rightarrow\PosX_s$ ray, similar to Eqn.~\ref{eqn:acacc}:
\begin{align}
    \PosX_s & = \PosX_t = \PosX_{t-1} -\LDir \Delta x = \LPos - t\LDir \Delta x, \label{eqn:xtacc} \\
    \accumAlpha_{\lt} & = \accumAlpha_{\ltp} + [1 - \accumAlpha_{\ltp}] \alpha(\PosX_t).\label{eqn:alacc}
\end{align}

In this rendering process (Eqn.~\ref{eqn:xsacc}-\ref{eqn:alacc}), a scene is represented by an opacity volume $\alpha$, a normal volume $\SNormal$ and a BRDF volume $\SReflect$; together, these express the geometry and reflectance of the scene, and we refer to them as \emph{Deep Reflectance Volumes}. 
The simplified opacity volume $\alpha$ is essentially one minus the transmission $\Trans$ (depending on the physical extinction coefficient $\Extinct$) over a ray segment of a fixed step size $\Delta x$; this means that $\alpha$ is dependent on $\Delta x$. 

Our physically-based ray marching is fully differentiable, so it can be easily incorporated in a deep learning framework and backpropagated through. With this rendering module, we present a neural rendering framework that simultaneously learns scene geometry and reflectance from captured images.

We support any differentiable reflectance model $f_r$ and, in practice, use the simplified Disney BRDF model \cite{karis2013real} that is parameterized by diffuse albedo and specular roughness (please refer to the supplementary materials for more details). 
Our opacity volume is a general geometry representation, accounting for both occlusions (view opacity accumulation in Eqn.~\ref{eqn:acacc})
and shadows (light opacity accumulation in Eqn.~\ref{eqn:alacc}).
We illustrate our neural rendering with ray marching in Fig.~\ref{fig:raymarching}.
Note that, because our acquisition setup has collocated camera and lighting,
 $\accumAlpha_{\lt}$ becomes equivalent to $\accumAlpha_{\cs}$ during training, thus requiring only one-pass opacity accumulation from the camera.
However, the learned opacity can still be used for re-rendering under any \emph{non-collocated lighting} with two-pass opacity accumulation.

Note that while alpha compositing-based rendering functions have been used in previous work on view synthesis, their formulations are not physically-based \cite{lombardi2019neural} and are simplified versions that don't model lighting \cite{srinivasan2019pushing,zhou2018stereo}. 
In contrast, our framework is physically-based and models single-bounce light transport with complex reflectance, occlusions and shadows.




\section{Learning Deep Reflectance Volumes}\label{sec:learning}

\subsection{Overview}
Given a set of images of a real scene captured under multiple known viewpoints with collocated lighting, we propose to use a neural network to reconstruct a Deep Reflectance Volume representation of a real scene. 
Similar to Lombardi et al. \cite{lombardi2019neural}, our network starts from a 512-channel deep encoding vector that encodes scene appearance; in contrast to their work, where this volume only represents RGB colors, we decode a vector to an opacity volume $\alpha$, normal volume $\SNormal$ and reflectance volume $\SReflect$ for rendering. 
Moreover, our scene encoding vector is not predicted by any network encoder; instead, we jointly optimize for a scene encoding vector and scene-dependent decoder network.

Our network infers the geometry and reflectance volumes in a transformed 3D space with a learned warping function $W$.
During training, our network learns the warping function $W$, and the geometry and reflectance volumes $\alpha_w$, $\SNormal_w$, $\SReflect_w$, where the subscript $w$ refers to a volume in the warped space.
The corresponding world-space scene representation is expressed by $V(\PosX) = V_w(W(\PosX))$, where $V$ is $\alpha$, $\SNormal$ or $\SReflect$.
In particular, we use bilinear interpolation to fetch a corresponding value at an arbitrary position $\PosX$ in the space from the discrete voxel values.
We propose a decoder-like network, which learns to decode the warping function and the volumes from the deep scene encoding vector.
We use a rendering loss between rendered and captured images as well as two regularizing terms.

\subsection{Network architecture}
\boldstartspace{Geometry and reflectance.} To decode the geometry and reflectance volumes  ($\alpha_w$, $\SNormal_w$, $\SReflect_w$),
we use upsampling 3D convolutional operations to 3D-upsample the deep scene encoding vector to a multi-channel volume that contains the opacity, normal and reflectance.
In particular, we use multiple transposed convolutional layers with stride 2 to upsample the volume, each of which is followed by a LeakyRelu activation layer.
The network regresses an 8-channel $128\times 128\times 128$ volume that includes $\alpha_w$, $\SNormal_w$ and $\SReflect_w$---one channel for opacity $\alpha_w$, three channels for normal $\SNormal_w$,
and four channels for reflectance $\SReflect_w$ (three for albedo and one for roughness).
These volumes express the scene geometry and reflectance in a transformed space, which can be warped to the world space for ray marching.

\boldstartspace{Warping function.}
To increase the effective resolution of the volume, we 
learn an affine-based warping function similar to~\cite{lombardi2019neural}.
The warping comprises a global warping and a spatially-varying warping.
The global warping is represented by an affine transformation matrix $W_g$.
The spatially varying warping is modeled in the inverse transformation space, 
which is represented by six basis affine matrices $\{W_j\}_{j=1}^{16}$
and a $32\times 32\times 32$ 16-channel volume $B$ that contains spatially-varying linear weights of the 16 basis matrices.
Specifically, given a world-space position $\PosX$, the complete warping function $W$ maps it into a transformed space by:
\begin{equation}
   W(\PosX) = [\sum_{j=1}^{16} B_j(\PosX)W_j]^{-1}W_g\PosX,
\end{equation}
where $B_j(\PosX)$ represents the normalized weight of the $j$th warping basis at $\PosX$.
Here, each global or local basis affine transformation matrix $W_*$ is composed of rotation, translation and scale parameters,
which are optimized during the training process.
Our network decodes the weight volume $B$ from the deep encoding vector using a multi-layer perceptron network with fully connected layers.

\subsection{Loss function and training details}
\label{sec:loss}
\boldstartspace{Loss function.}
Our network learns the scene volumes using a rendering loss computed using the differentiable ray marching process discussed in Sec.~\ref{sec:volume}. 
During training, we randomly sample pixels from the captured images and do the ray marching (using known camera calibration) to get the rendered pixel colors $\Radiance_k$ of pixel $k$;
we supervise them with the ground truth colors $\tilde{\Radiance}_k$ in the captured images using a $L_2$ loss.
In addition, we also apply regularization terms from additional priors similar to \cite{lombardi2019neural}.
We only consider opaque objects in this work and enforce the accumulated opacity along any camera ray $\accumAlpha_{c_k \rightarrow s'}$ (see Eqn.~\ref{eqn:acacc}, here $k$ denotes a pixel and $s'$ reflects the final step that exits the volume) to be either 0 or 1, corresponding to a background or foreground pixel, respectively. 
We also regularize the per-voxel opacity to be sparse over the space by minimizing the spatial gradients of the logarithmic opacity.
Our total loss function is given by:
\begin{equation}
    \sum_k\|\Radiance_k-\tilde{\Radiance}_k\|^2 + \beta_1\sum_k[\log(\accumAlpha_{c_k \rightarrow s'})+\log(1-\accumAlpha_{c_k \rightarrow s'})]
    +\beta_2\sum \| \nabla_{\PosX} \log \alpha(\PosX)\|
    \label{eqn:loss}
\end{equation}
Here, the first part reflects the data term, the second regularizes the accumulated $\alpha$ and the third regularizes the spatial sparsity.

\boldstartspace{Training details.}
We build our volume as a cube located at $[-1,1]^3$.
During training, we randomly sample $128\times 128$ pixels from $8$ captured 
images for each training batch, and perform ray marching through the volume 
using a step size of $1/64$.
Initially, we set $\beta_1=\beta_2=0.01$;
we increase these weights to $\beta_1=1.0$, $\beta_2=0.1$ after $300000$ iterations, which helps remove the artifacts in the background and 
recover sharp boundaries.


\section{Results}\label{sec:results}
In this section we show our results on real captured scenes. We first introduce our acquisition 
setup and data pre-processing.  Then we compare against the
state-of-the-art mesh-based appearance acquisition method, followed by a detailed analysis  
of the experiments. We also demonstrate material editing results with our approach. Please 
refer to the supplementary materials for video results.

\begin{figure}[t]
     \includegraphics[width=\textwidth]{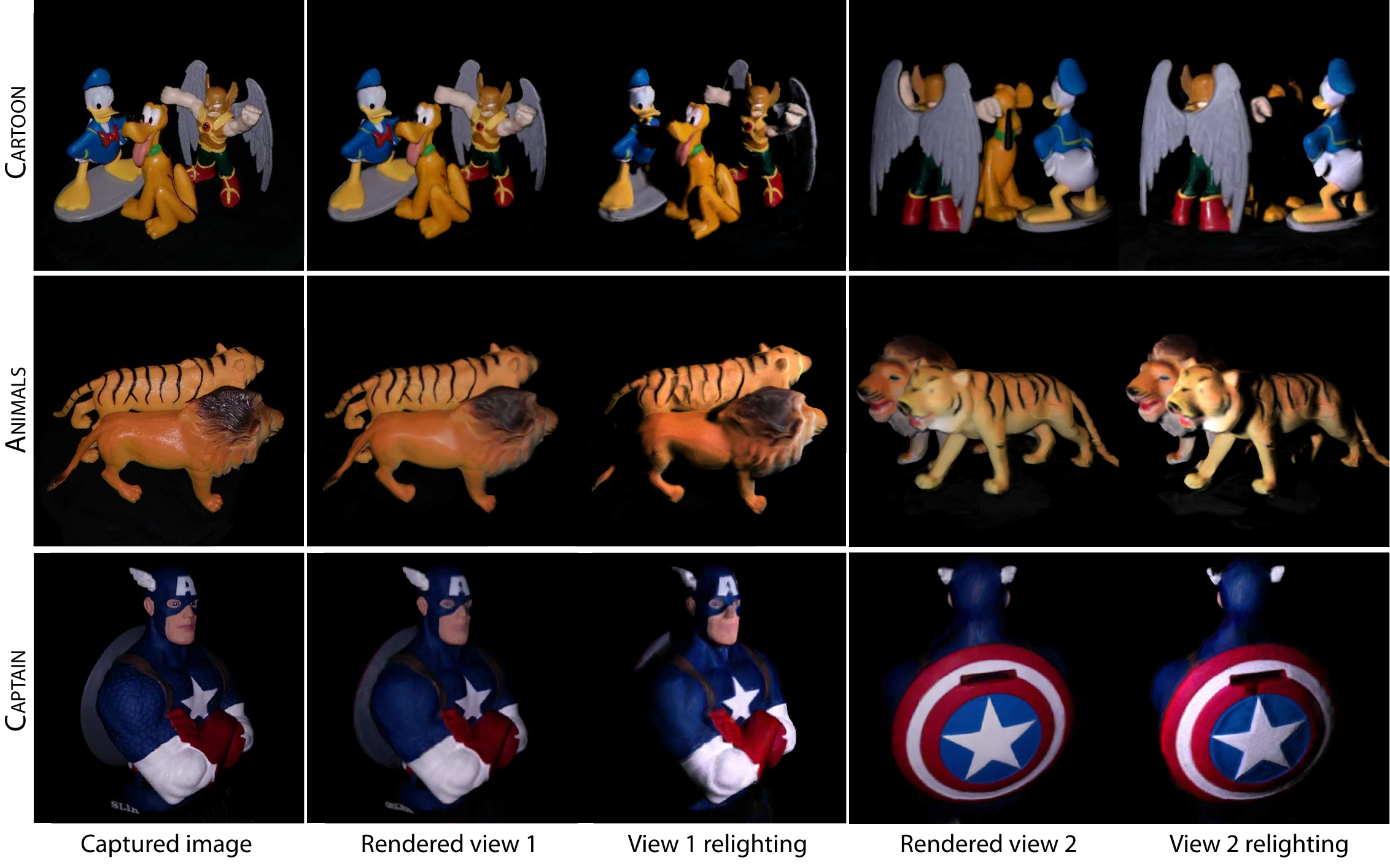}
    \caption{
       Additional results on real scenes. We show renderings under novel
        view and lighting conditions. Our method is able to handle scenes 
        with multiple objects (top two rows) and model the complex occlusions between them.
       Our method can also generate high-quality results from casual handheld video captures (third row), which demonstrates the 
       practicability of our approach. }
    \label{fig:results}
\end{figure}

\boldstartspace{Data acquisition.}
Our approach learns the volume representation in a scene dependent way from images with collocated view and light; 
this requires adequately dense input images well distributed around a target scene to learn complete appearance.
Such data can be practically acquired by shooting a video using 
a handheld cellphone; we show one result using this practical handheld setup in Fig.~\ref{fig:results}. 
For other results, we use a robotic arm to automatically capture more uniformly distributed images around scenes
for convenience and thorough evaluations;
this allows us to evaluate the performance of our method with different numbers of input images that
are roughly uniformly distributed as shown in Tab.~\ref{tab:inputnumber}.  In the robotic arm
setups, we mount a Samsung Galaxy Note 8 cellphone to the robotic arm and capture about 480 images
using its camera and the built-in flashlight in a dark room; we leave out a subset of 100 images 
for validation purposes and use the others for training. We use the same phone to capture 
a $4$-minute video of the object in \SCaptain~and select one image for training for every $20$ frames, which effectively gives us $310$ training images.

\boldstartspace{Data pre-processing.}
Our captured objects are roughly located around the center of the images. 
We select one fixed rectangular region around the center that covers the object across all frames and use it to crop the images as input for training.
The resolution of the cropped training images fed to our network ranges from $400 \times 500$ to $1100 \times 1100$.
Note that we do not use a foreground mask for the object.
Our method leverages the regularization terms in training (see Sec.~\ref{sec:loss}), which automatically recovers a clean background.
We calibrate the captured images using structure from motion
(SfM) in COLMAP~\cite{schoenberger2016sfm}
to get the camera intrinsic and extrinsic parameters.
Since SfM may fail to register certain views, the actual number of training images varies from 300 to 385 in different scenes. We estimate 
the center and bounding box of the captured object with the sparse reconstructions from SfM. We translate the center of the object to the origin 
and scale it to fit into the $[-1, 1]^3$ cube. 

\begin{figure}[t]
    \centering
    \begin{minipage}{0.5\textwidth}
        \centering
        \includegraphics[width=\textwidth]{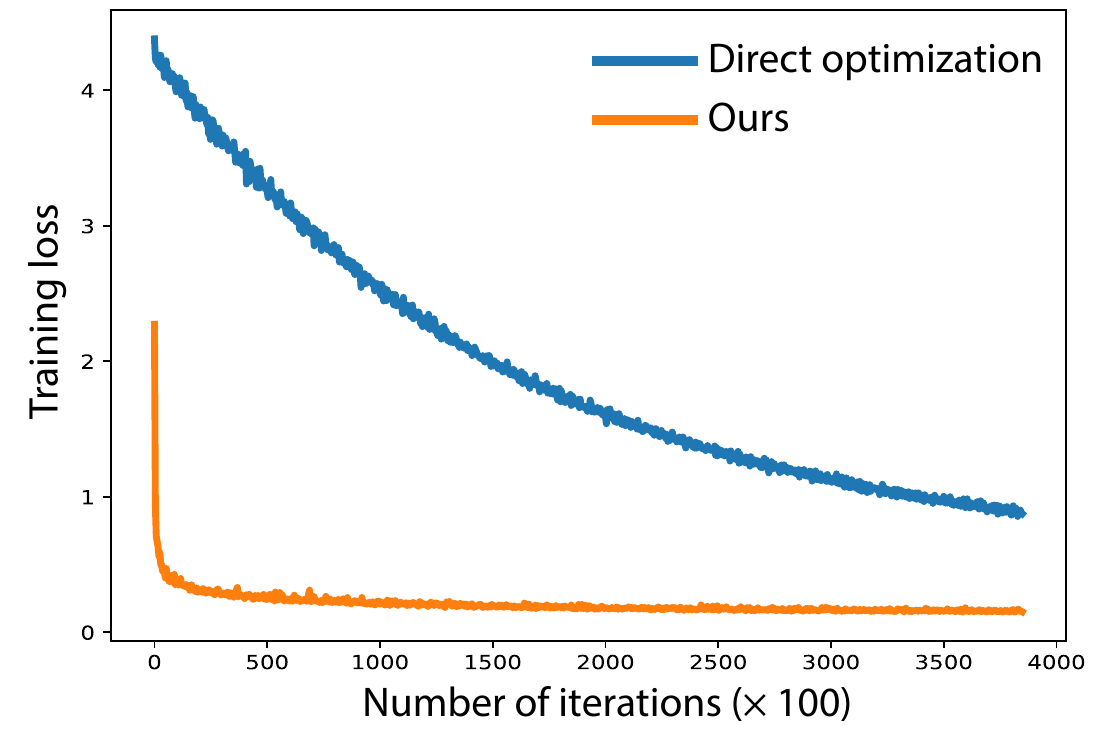}
        \caption{
            We compare our deep prior based optimization against direct
            optimization of the volume and warping function without using
            networks. Direct optimization converges significantly slower
            than our method, which demonstrates the effectiveness of 
            regularization by the networks.
        }
        \label{fig:optimization}
    \end{minipage}\hfill
    \begin{minipage}{0.47\textwidth}
        \centering
        \includegraphics[width=\textwidth]{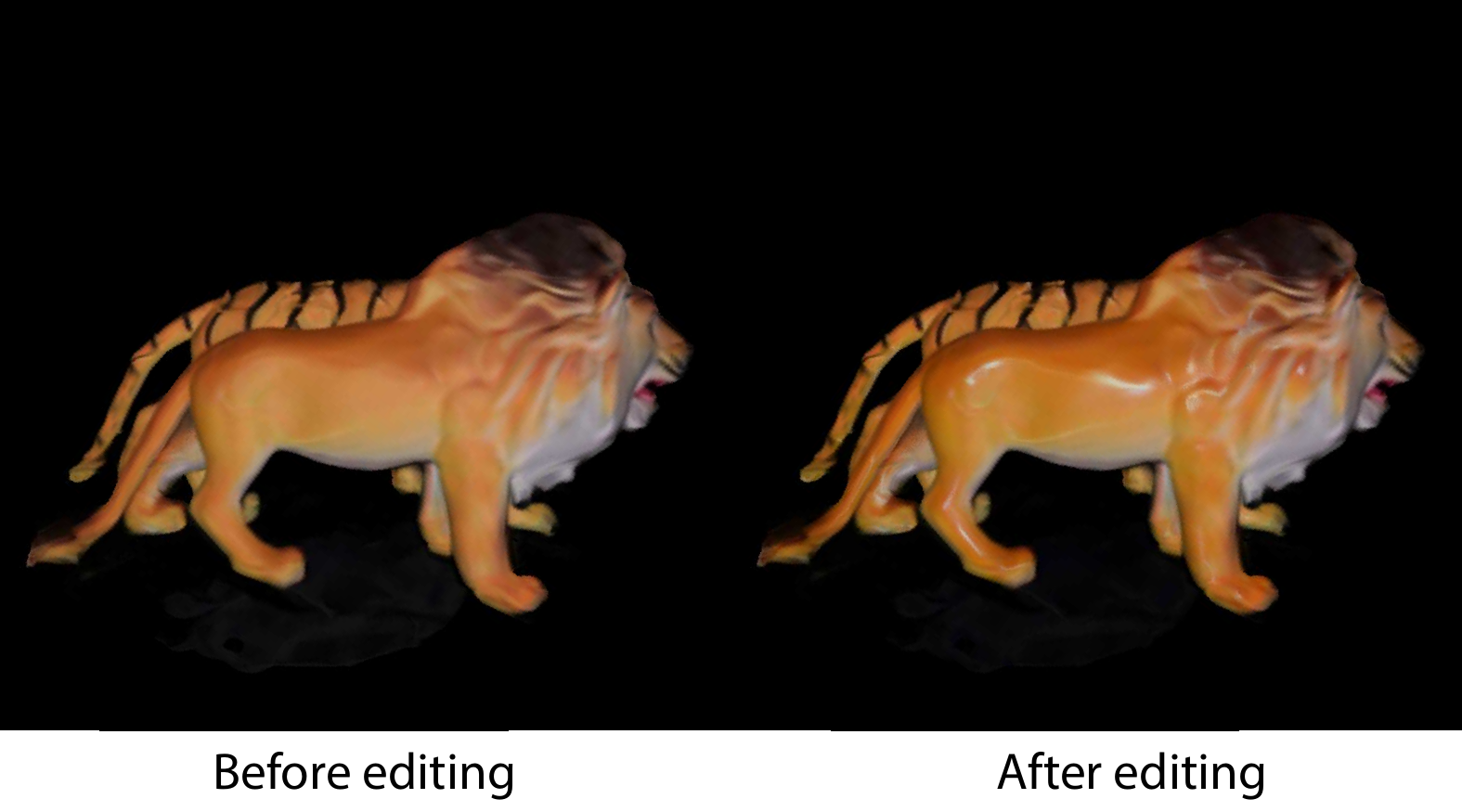}
        \caption{
            Our approach supports intuitive editing of the material
            properties of a captured object. In this example we decrease the
            roughness of the object to make it look like glossy marble
            instead of plastic.
        }
        \label{fig:material}
    \end{minipage}
\end{figure}

\boldstartspace{Implementation and timing.}
We implement our system (both neural network and differentiable volume rendering components) using PyTorch. 
We train our network using four NVIDIA 2080Ti RTX GPUs for about two days (about 450000 iterations; though 200000 iterations for 1 day typically already converges to good results, see Fig.~\ref{fig:optimization}).
At inference time, we directly render the scene from the reconstructed volumes without the network.
It takes about 0.8s to render a $700\times 700$ image 
under collocated view and light.
For non-collocated view and light, the rendering requires connecting each shading point to the light source with additional light-dependent opacity accumulation,
which is very expensive if done naively.
To facilitate this process, we perform ray marching
from the light's point of view and precompute the
accumulated opacity at each spatial position of the volume. 
During rendering, the accumulated opacity for the light ray 
can be directly sampled from the precomputed volume. 
By doing so, our final rendering under arbitrary light and view takes about $2.3$s.

\begin{figure}[h]
    \includegraphics[width=\textwidth]{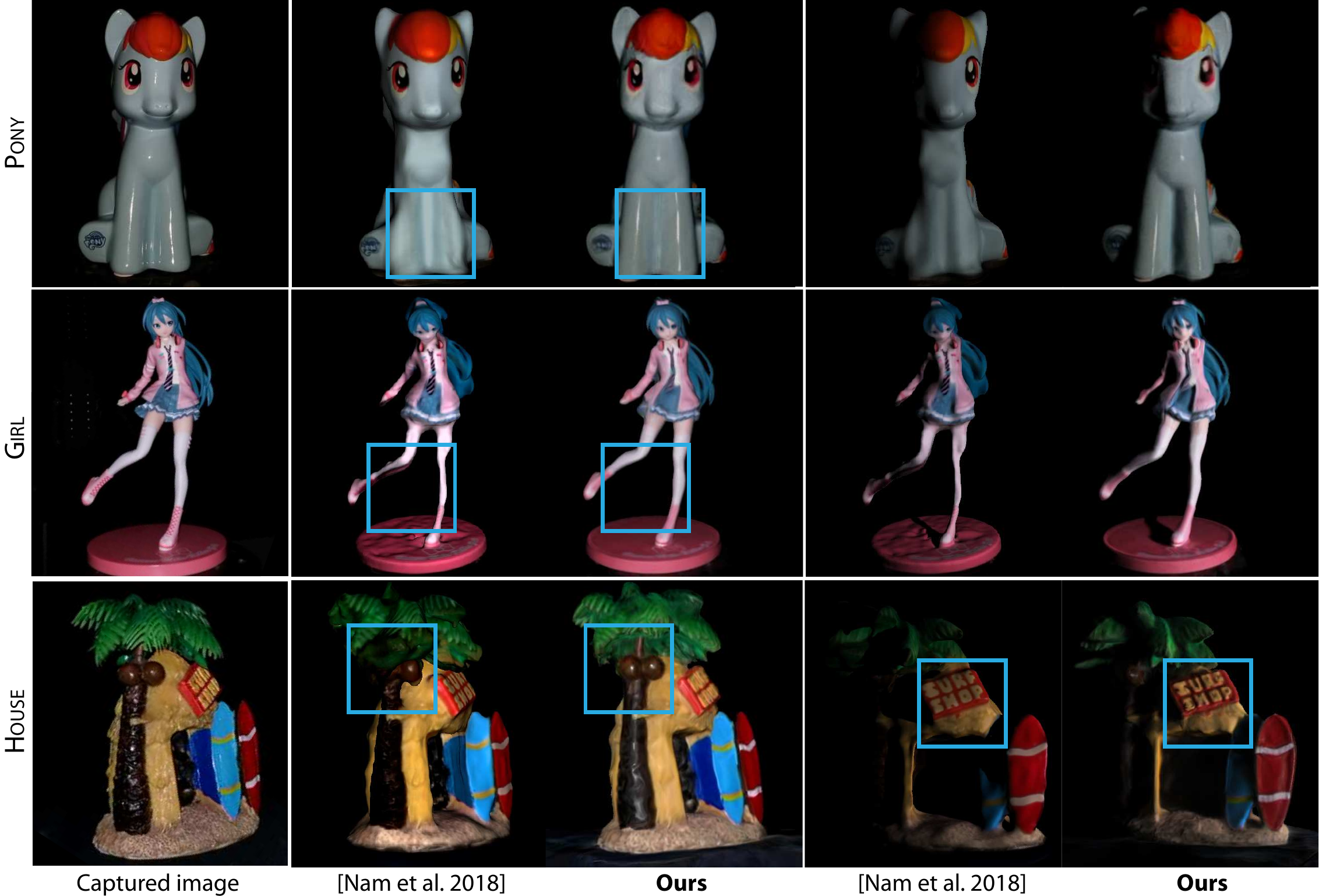}
    \caption{Comparisons with mesh-based reconstruction. 
    We show renderings of the captured object under both 
    collocated (column 2, 3) and non-collocated (column 4, 5)
    camera and light.  
    We compare our volume-based neural reconstruction against a state-of-the-art
    method \cite{nam2018practical} that reconstructs mesh and 
    per-vertex BRDFs. Nam et al.~\cite{nam2018practical} fails to handle such 
    challenging cases and recovers inaccurate geometry and appearance.
    In contrast our method produces photo-realistic results. 
    }
    \label{fig:meshbased}
\end{figure}
\boldstartspace{Comparisons with mesh-based reconstruction.}
We use a practical acquisition setup where we capture 
unstructured images using a mobile phone 
with its built-in flashlight on in a dark room.
Such a mildly controlled acquisition setup is rarely 
supported by previous
works~\cite{bi2020deep,kang2019learning,xia2016recovering,xu2019deep,xu2018deep,zhou2016sparse}. 
Therefore, we compare with the state-of-the-art method 
proposed by Nam et al.~ \cite{nam2018practical} for mesh-based geometry 
and reflectance reconstruction, that uses the same cellphone setup as ours 
to reconstruct a mesh with per-vertex BRDFs, and supports both relighting and view synthesis.
Figure~\ref{fig:meshbased} shows comparisons on renderings under both collocated and
non-collocated view-light conditions.
The comparison results are generated from the same set of input images, 
and we requested the authors of~\cite{nam2018practical} run their code on our data and compared on the rendered 
images provided by 
the authors. Please refer to the supplementary materials for video 
comparisons.


As shown in Fig.~\ref{fig:meshbased}, our results are significantly better than the mesh-based method in terms of both geometry and reflectance.
Note that, Nam et al.~\cite{nam2018practical} leverage a state-of-the-art MVS method \cite{schoenberger2016mvs} to reconstruct
the initial mesh from captured images and performs an optimization to further refine the geometry;
this however still fails to recover the accurate geometry in texture-less, specular and thin-structured regions in those challenging scenes,
which leads to seriously distorted shapes in \SPony, over-smoothness and undesired structures in \SHouse, 
and degraded geometry in \SGirl.
Our learning-based volumetric representation avoids these mesh-based issues and models the scene geometry accurately with many details.
Moreover, it is also very difficult for the classical per-vertex BRDF optimization in \cite{nam2018practical} 
to recover high-frequency specularities, which leads to over-diffuse appearance in most of the scenes;
this is caused by the lack of constraints for the high-frequency specular effects, 
which appear in very few pixels in limited input views.
In contrast, our optimization is driven by our novel neural rendering framework with deep network priors, 
which effectively correlates the sparse specularities in different regions through network
connections and recovers realistic specularities and other appearance effects.

\begin{figure}[h!!]
    \centering
    \begin{minipage}{0.49\textwidth}
        \centering
        \setlength{\tabcolsep}{3.0pt}
        \begin{tabular}{crrrrr}
            \hline
            & 25 & 50 & 100 & 200 & 385 \\ \hhline{======}
            PSNR & 25.33 &  26.36 & 26.95 &  27.85 &  28.13 \\  
            SSIM & 0.70 & 0.73 & 0.75 &  0.80 & 0.81\\  \hline 
        \end{tabular}
        \caption{
        We evaluate the performance of our method on the \SHouse~scene with
        different numbers of training images. Although we use all $385$
        images in our final experiments, our method is able to achieve
        comparable performance with as few as 200 images for this challenging
        scene.
        }
    \label{tab:inputnumber}
    \end{minipage}\hfill
    \begin{minipage}{0.50\textwidth}
        \centering
        \setlength{\tabcolsep}{4.0pt}
        \begin{tabular}{crr}
            \hline
            & \textsc{House} & \textsc{Cartoon} \\ \hhline{===}
            ~\cite{sitzmann2019deepvoxels} & 0.786/25.81 &   0.532/16.34 \\ 
            \textbf{Ours} & \textbf{0.896}/\textbf{30.44} & 
            \textbf{0.911}/\textbf{29.14} \\ \hline \end{tabular}
        \caption{
            We compare against DeepVoxels on 
            synthesizing novel views under collocated lights and report 
            the PSNR/SSIM scores. The results show that our method generates
            more accurate renderings. Note that we retrain our model  
            with a resolution of $512 \times 512$ for a fair 
            comparison.
        }
        \label{tab:comp_dv}
    \end{minipage}
\end{figure}

\boldstartspace{Comparison on synthesizing novel views.}
    We also make a comparison on synthesizing 
    novel views under collocated lights against a view synthesis method DeepVoxels~\cite{sitzmann2019deepvoxels}, which encodes view-dependent
    appearance in a learnt 3D-aware neural representation.
    Note that DeepVoxels does not support relighting. 
    As shown in Fig.~\ref{tab:comp_dv}, our method is able 
    to generate renderings of higher quality with higher PSNR/SSIM scores. 
    In contrast, DeepVoxels fails to reason about the 
    complex geometry
    in our real scenes, thus resulting in degraded image quality.
    Please refer to the supplementary materials for visual comparison results.

\boldstartspace{Additional results.}
We show additional relighting and view synthesis results of complex real scenes in Fig.~\ref{fig:results}.
Our method is able to handle scenes with multiple objects, as shown in 
scene \textsc{Cartoon}~and \SLion. Our volumetric representation can accurately model 
complex occlusions between objects and reproduce realistic cast shadows under novel lighting,
which are never observed by our network during the training process. 
In the \SCaptain~scene, we show the result generated from \emph{handheld} mobile phone captures. 
We select frames from the video at fixed intervals as training data. Despite the potential 
existence of motion blur and non-uniform coverage, our method is able to generate high-quality 
results, which demonstrates the robustness and practicality of our approach. Please refer to the supplementary materials for video results.


\boldstartspace{Evaluation of the number of inputs.}
Our method relies on an optimization over adequate input images that capture the scene appearance 
across different view/light directions.
We evaluate how our reconstruction degrades with the decrease of training images on the
\SHouse~scene. 
We uniformly select a subset of views from the full training images
and train our model on them.  
We evaluate the trained model on the test images, and report the SSIMs and PSNRs in
Fig.~\ref{tab:inputnumber}. 
As we can see from the results, there is an obvious performance 
drop when there are fewer than 100 training images due to insufficient constraints.
On the other hand, while we use the full $385$ images for our final results, our method in fact achieves comparable performance with only $200$ for this scene, as reflected by their close PSNRs and SSIMs.

\boldstartspace{Comparison with direct optimization.}
Our neural rendering leverages a ``deep volume prior'' to drive the volumetric optimization process.
To justify the effectiveness of this design, 
we compare with a naive method that directly optimizes the parameters in each voxel and the warping 
parameters using the same loss function.
We show the optimization progress in Fig.~\ref{fig:optimization}.
Note that, the naive method converges significantly slower than ours, where 
the independent voxel-wise optimization without considering across-voxel correlations
cannot properly disentangle the ambiguous information in the captured images;
yet, our deep optimization is able to correlate appearance information across the voxels
with deep convolutions, which effectively minimizes the reconstruction loss.


\boldstartspace{Material editing.}
Our method learns explicit volumes with physical meaning to represent the reflectance of real scenes.
This enables broad image synthesis applications like editing the materials of captured scenes.
We show one example in Fig.~\ref{fig:material}, 
where we successfully make the scene glossier by decreasing the learned roughness in the volume.
Note that, the geometry and colors are still preserved in the scene, 
while novel specularities are introduced which are not part of the material appearance in the scene.
This example illustrates that our network disentangles the geometry and reflectance of the scene in a reasonable way,
thereby enabling sub-scene component editing without influencing other components.


\boldstartspace{Limitations.} 
We reconstruct the deep reflectance volumes with a resolution of
$128^3$, which is restricted by available GPU memory.
While we have applied a warping function to increase the actual utilization of the volume space,
and demonstrated that it is able to generate compelling results on complex real scenes, 
it may fail to 
fully reproduce the geometry and appearance of scenes with highly complex surface normal
variations and texture details.
Increasing the volume resolution may resolve this issue.
In the future, it would also be interesting to 
investigate how to efficiently apply sparse representations such as octrees in our framework to increase the 
capacity of our volume representation. 
The current reflectance model we are using is most appropriate for opaque
surfaces. Extensions to other materials like hair, fur or glass could be
potentially addressed by applying other reflectance models in our neural
rendering framework.



\section{Conclusion}\label{sec:conclusions}
We have presented a novel approach to learn a volume 
representation that models both geometry and reflectance of complex real scenes. 
We predict per-voxel opacity, 
normal, and reflectance from unstructured multi-view mobile phone captures
with the flashlight. 
We also introduce a physically-based differentiable rendering module to 
enable renderings of the volume under arbitrary viewing and lighting directions. 
Our method is practical, and supports novel view synthesis,
relighting and material editing, which has significant potential benefits
in scenarios such as 3D visualization and VR/AR applications. 

\boldstartspace{Acknowledgements.}
We thank Giljoo Nam for help with the comparisons. This work was supported in
part by ONR grants N000141712687, N000141912293, N000142012529, NSF grant
1617234, Adobe, the Ronald L. Graham Chair and the UC San Diego Center for Visual Computing.

{
\bibliographystyle{splncs04}
\bibliography{egbib}
}

\appendix
\section{BRDF Model}\label{sec:brdf}
Essentially any differentiable BRDF model can be incorporated in our framework 
to model the appearance of real-world objects. In this paper we apply a  version of the microfacet BRDF model proposed by Walter et al.~\cite{walter2007microfacet}, with simplifications introduced by Karis~\cite{karis2013real}.
Let $\EDir$, $\LDir$ be the view and light direction, $\SNormal$, $\SAlbedo$, $\SRough$ be the 
normal, diffuse albedo and roughness. 
Our BRDF model is defined as:
\begin{equation}
f(\EDir, \LDir,\SAlbedo, \SNormal, \SRough) = \frac{\SAlbedo}{\pi} + \frac{D(\HDir, \SNormal, r) F(\EDir,
\HDir)G(\LDir, \EDir, r)}{4(\SNormal \cdot \LDir)(\SNormal \cdot \EDir)}
\end{equation}
where $D(\HDir, \SNormal, r)$, $F(\EDir, \HDir)$ and $G(\LDir, \EDir, \HDir, r)$ are the 
\emph{normal distribution}, \emph{fresnel} and \emph{geometric terms} respectively. These terms are defined as follows:
\begin{eqnarray}
D(\HDir, \SNormal, \SRough) &=& \frac{\alpha^{2}}{\pi\left[(\SNormal\cdot \HDir)^{2}(\alpha^{2} - 1) + 1\right]^{2} } \nonumber  \\
\alpha &=& \SRough ^{2} \nonumber \\
F(\EDir, \HDir) &=& F_0 + (1 - F_0 ) 2^{ -\left[5.55473(\EDir \cdot \HDir) + 6.8316\right](\EDir \cdot \HDir)} \nonumber \\
G(\LDir, \EDir, \SRough) &=& G_{1}(\EDir, \SNormal)G_{1}(\LDir, \SNormal) \nonumber \\ 
G_{1}(\EDir, \SNormal) &=& \frac{\SNormal \cdot \EDir}{(\SNormal \cdot \EDir)(1 -k) + k} \nonumber \\
G_{1}(\LDir, \SNormal) &=& \frac{\SNormal  \cdot \LDir}{(\SNormal  \cdot \LDir)(1 -k) + k} \nonumber \\
k &=& \frac{(\SRough + 1)^2}{8} \nonumber 
\end{eqnarray}
where we set $F_0 = 0.05$ as suggested in~\cite{karis2013real}. 
Correspondingly, the final reflected radiance $f_r$ in Eqn.~8 in the paper is computed as:
\begin{equation}
    f_r(\EDir,\LDir,\SNormal(\PosX_s),\SReflect(\PosX_s)) = f(\EDir,\LDir,  \SAlbedo(\PosX_s), \SNormal(\PosX_s), \SRough(\PosX_s)) (\SNormal(\PosX_s) \cdot \LDir)
\end{equation}
where $\SAlbedo(\PosX_s)$ and $\SRough (\PosX_s)$ are the diffuse albedo and  roughness at $\PosX_s$. 

\section{Network Architecture}\label{sec:network}
\begin{figure}[t]
    \includegraphics[width=\textwidth]{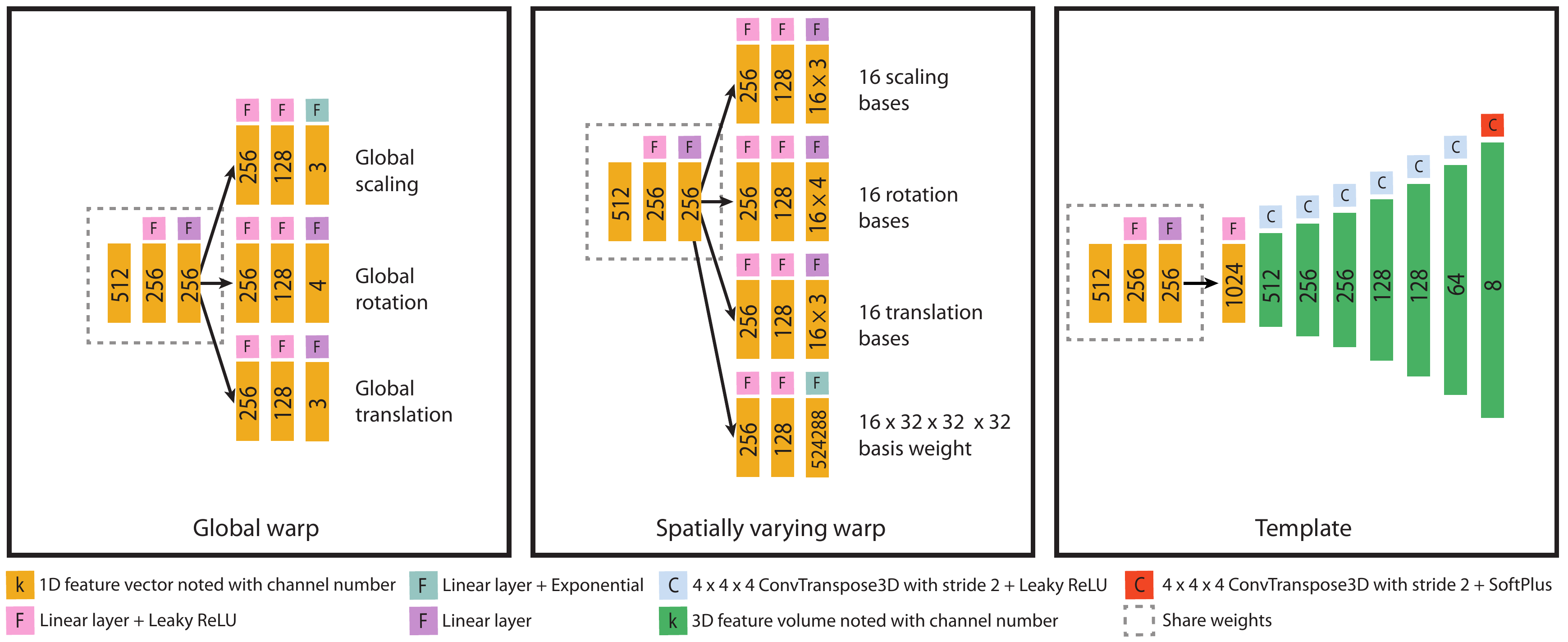}
    \caption{Our network architecture.}
    \label{fig:network}
\end{figure}

Fig.~\ref{fig:network} shows an overview of our network architecture.  Our network starts from a 
$512$-channel encoding vector initialized using random samples from a normal distribution. The
encoding vector first goes through two fully connected layers and then is fed to different decoders 
to predict the global warping parameters, 
spatially varying warping parameters, and the template volume. The global warping parameters 
$W_g$ consist of a 3-channel scaling vector, a 3-channel translation vector and a 4-channel
rotation vector represented as a quaternion. The spatially varying parameters consist of 
$16$ warping bases $\{W_j\}_{j=1}^{16}$ and a weight volume $B$. Similar to the global warping, 
each warping basis is composed of a scaling, a translation and a rotation. The weight volume $B$
has $16$ channels and a resolution of $32\times32\times32$, which encodes the spatially varying 
weight of each basis. Finally, the template volume $V$ has a resolution of $128 \times 128 \times
128$; it has $8$ channels with 1 channel for opacity, 3 channels for normal, 3 channels for 
diffuse albedo  and $1$ channel for roughness. We also transform the albedo and 
roughness to the range of $[0, 1]$ and normalize the predicted normal vectors.

\begin{table}[t]
        \centering
        \renewcommand{\arraystretch}{1.2}
        \setlength{\tabcolsep}{6.0pt}
        \begin{tabular}{lcccccc}
            \hline
             & \SPony & \SGirl & \SHouse & \SDisney & \SLion &  \SCaptain \\ \hhline{=======}
             min & 0.60  & 0.82 & 1.22 & 0.25 & 0.29 &  0.68\\ 
             max & 10.00  &  9.19 & 9.54 & 10.09 & 9.28 & 14.52 \\ 
             mean & 5.35  &  7.66 & 5.83 & 7.25 & 6.49 & 6.92\\ \hline
        \end{tabular}

        \caption{
        The minimum, maximum and average angles (in degrees) between the test views in the supplementary 
        video and  their nearest training views. 
    }\label{tab:angle}

\end{table}

\section{Testing Specifications}\label{sec:testing}
In the supplementary video, we show renderings of the captured object under novel viewpoints and 
lighting. Note that our training images are captured with collocated light and camera, and 
the relighting results in the video demonstrate that our volumetric representation can generalize 
to novel lighting conditions. In Tab.~\ref{tab:angle}, we report the minimum, maximum and average 
angles between the test views in the video and their nearest training views. Such a large angle difference 
also shows that our deep reflectance volume generalizes well to novel views.

\begin{figure}[t]
    \includegraphics[width=\textwidth]{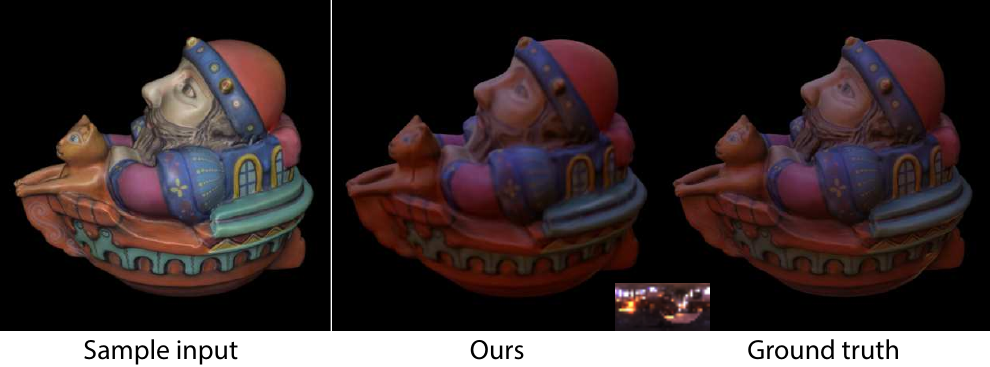}
    \caption{
           Comparison with ground truth on relighting under environment illumination. The environment map used 
           for rendering is shown at the bottom.
    }
    \label{fig:comp_envmap}
\end{figure}

\begin{figure}[t]
    \includegraphics[width=\textwidth]{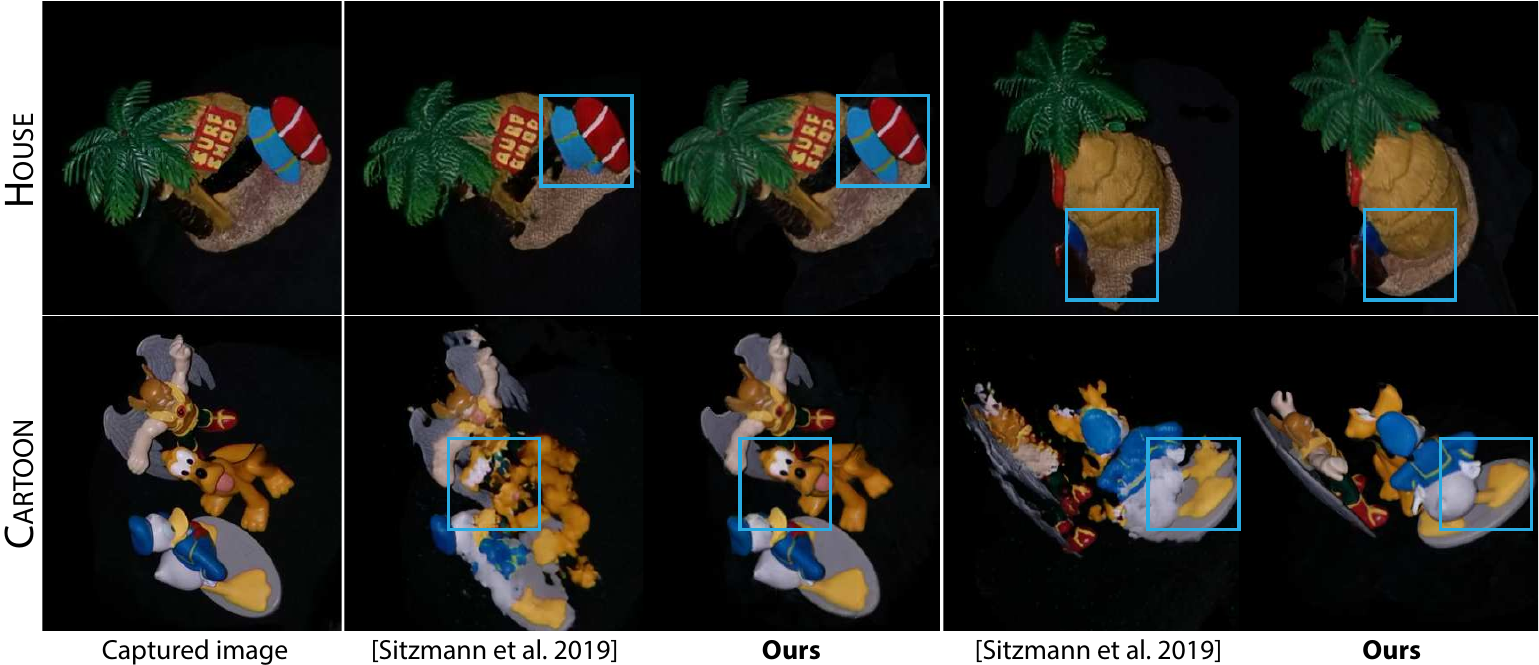}
    \caption{
        Comparison against Sitzmann et al.~\cite{sitzmann2019deepvoxels}
        on synthesizing novel views under collocated lights.
        Our method is able to generate high-quality results with fewer artifacts.
    }
    \label{fig:comp_dv}
\end{figure}

\begin{figure}[t]
    \includegraphics[width=\textwidth]{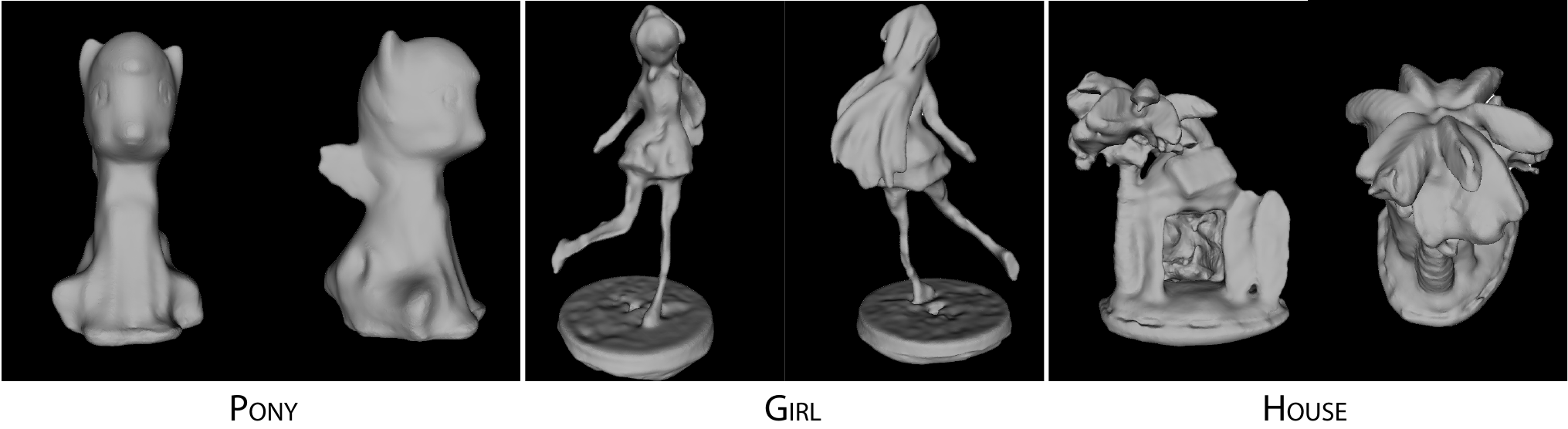}
    \caption{Geometry reconstructed from Nam et al.~\cite{nam2018practical}. }
    \label{fig:geo}
\end{figure}

\section{Results on Synthetic Data}
In addition to the real captures, we also evaluate 
our method on a synthetic dataset where we render a 
synthetic scene from multiple viewpoints under 
collocated camera and light.
We compare our view synthesis and relighting results 
with the ground truth renderings. 
Please check the supplementary video for comparisons. 

By linearly combining the relit images under each light corresponding to pixels of an environment map, our method 
also supports rendering of the scene under novel environment illumination. In Fig.~\ref{fig:comp_envmap} we demonstrate 
our environment map relighting result and compare it to the ground 
truth renderings with a physically-based renderer. From the figure 
we can see that our method can generate visually plausible
results.

\section{Comparison on View Synthesis}
    In Fig.~\ref{fig:comp_dv} we show a visual comparison
    against the method proposed by Sitzmann et al.~\cite{sitzmann2019deepvoxels} on synthesizing novel
    views under collocated lights. 
    Sitzmann et al. learn a 3D-aware neural representation to encode the view-dependent appearance of 
    captured scenes. Their method cannot model the complex geometry and appearance of our real scenes. As we can see from the result, Sitzmann et al. cannot synthesize novel views 
    correctly and  generates distorted images with undesired structures. In contrast, our method is able to produce 
    images of much higher quality.

\section{Mesh-Based Appearance Acquisition}\label{sec:geo}
In Fig.~\ref{fig:geo} we show the optimized geometry from Nam et al.~\cite{nam2018practical}. They 
leverage the state-of-the-art multi-view stereo (MVS) framework to get an initial geometry, and further perform
an optimization to refine it; however they still fail to recover the faithful geometry for such challenging 
scenes where there are textureless and thin-structured regions, thus resulting in degraded quality in reproduced
appearance, as shown in the supplementary video.

\end{document}